# Exploiting Uniform Assignments in First-Order MPE

**Udi Apsel** and **Ronen I. Brafman**
Computer Science Dept.
Ben-Gurion University of The Negev
Beer-Sheva, Israel 84105
apsel,brafman@cs.bgu.ac.il

## Abstract

The *MPE* (Most Probable Explanation) query plays an important role in probabilistic inference. MPE solution algorithms for probabilistic relational models essentially adapt existing belief assessment method, replacing summation with maximization. But the rich structure and symmetries captured by relational models together with the properties of the maximization operator offer an opportunity for additional simplification with potentially significant computational ramifications. Specifically, these models often have groups of variables that define symmetric distributions over some population of formulas. The maximizing choice for different elements of this group is the *same*. If we can realize this ahead of time, we can significantly reduce the size of the model by eliminating a potentially significant portion of random variables. This paper defines the notion of *uniformly assigned* and *partially uniformly assigned* sets of variables, shows how one can recognize these sets efficiently, and how the model can be greatly simplified once we recognize them, with little computational effort. We demonstrate the effectiveness of these ideas empirically on a number of models.

## 1 Introduction

Probabilistic relational models (PRM) [5] such as the *Markov Logic Network* (MLN) [13] and the *Parfactor model* [11] encapsulate a large number of random variables and potential functions using first-order predicate logic. Using exact lifted inference methods [2; 9; 12; 6; 4], one can perform inference tasks directly on the relational model, solving many tasks which were previously considered intractable.

To date, the first-order MPE query is perceived as a derivative of these methods, since its computational structure is similar to computing the partition function of a first-order model, with the exception of maxing-out random variables instead of summing-out, as laid out by Braz et al. [3].

In this paper we introduce a method that simplifies many first-order MPE tasks by utilizing properties specific to the maximization operator, allowing us to solve models in which computing the partition function is complicated, or even intractable.

The MPE property which we capitalize on is called a *uniform assignment* (UA). Briefly, it is the existence of a group of random variables that are assigned identically in some MPE solution. In relational models, these groups correspond to sets of ground atoms derived from the same atomic formula. Once a group is recognized as uniformly assigned, it can be replaced by a single representative, substantially simplifying the model. In relational models, this replacement corresponds to an arity reduction – the removal of some logical variable(s) from the relevant atomic formula. With suitable care, the MPE probability in the new model is the same as that of the original model, and the MPE for the original model is easily derived from the MPE of the new model.

The key contribution of this paper is a computationally efficient procedure for finding a set of uniform assignments in a given model. For this purpose, we define a set of purely symbolic operators (*anchoring*, *model alignment*, and *symbolic fusion*), which are all based on standard lifted inference tools, *fusion* [2] and *propositionalization* [9]. When applied repeatedly, the three symbolic operators serve as a structure analysis tool, and result in a compact representation of the UA property, expressing which logical variables can be removed from which atomic formulas.

Once uniform assignments are detected, we proceed to simplify our model by applying a reduction procedure called *uniform assignment reduction* (UAR). The procedure reduces the arity of relevant atomic formulas, and suitably modifies table entries in the model, ensuring that the MPE probability of the modified model remains equal to that of the original.

Both steps, UA detection and UA reduction, are preprocessing steps that often result in a much smaller model. Both incur low computational overhead that is independent of domain sizes, and are totally agnostic to the MPE algorithm later applied. As we demonstrate, the effort of UAR reduction is small w.r.t. to the complexity of the underlying inference task, and incurs virtually no overhead even when no simplification is achieved. Finally, the smaller, simplified, model is passed to your favorite MPE engine, and from the computed MPE, the MPE of the original model is easily extracted.

Our UA detection method is general, and can be applied both on MLNs and parfactor models. However, UA detection is most naturally formulated on parfactors, since it utilizes the (symbolic) fusion operator, which is not commonly applied to MLNs. We therefore start with a background of the parfactor model, before introducing the concepts of our method in a Walk-through section. Formal definitions and complete algorithm formulation are presented next, followed by an extension of UA to recursive conditioning, a presentation of the experimental results, and a conclusion section.

## 2 Background

We review the properties of the parfactor model, as introduced by Poole [11], and later extended by Braz et al. [2] and Milch et al. [9]. Readers familiar with this model can safely skip to the following Walk-through section.

### 2.1 Model Representation

The parfactor model is a first-order representation of a factor network, such as a Bayesian [10] or a Markov [13] network. Random variables in the model correspond to ground atomic formulas (aka *ground atoms*) of the form $p(c_1, \ldots, c_n)$, where $p$ is a predicate with an assignment range $range(p)$, and $c_1, \ldots, c_n$ are constant symbols. Under a set of assignments $v$ to random variables, the notation $\alpha_{|v}$ is used to depict the values assigned to $\alpha$, whether $\alpha$ is a ground atom or a set of ground atoms. A *factor* $f$ is a pair $(A, \phi)$, consisting of a set of ground atoms $A = \bigcup_{i=1}^{m} \{\alpha_i\}$ and a potential function $\phi : \prod_{i=1}^{m} range(\alpha_i) \to \mathbb{R}_0^+$. Under the set of assignments $v$, the weight of factor $f$ is $w_f(v) = \phi(\alpha_{1|v}, \ldots, \alpha_{m|v})$. The abbreviation $\phi : \alpha_i, \ldots, \alpha_m$ is used to denote a factor with a potential table $\phi$ and a respective set of ground atoms.

**Example 2.1.** Let factor $f$ consist of the following, $\phi : smokes(Alice), drinks(Bob), friends(Alice, Bob)$. Hence, the table entries in $\phi$ represent the various chances of Alice and Bob being friends, depending on whether Alice and Bob are drinkers/smokers.

Whereas factor networks are modeled by a set of factors, the core representation unit of our model is the parameterized factor (aka *parfactor*), a first-order extension of a factor. Before reviewing its properties, we define the notions of *atoms*, *logical variables*, *substitutions* and *constraints*.

An atomic formula (aka *atom*) is a formula of the form $p(t_1, \ldots, t_n)$, where $t_i$ is a constant or a *logical variable*. Each logical variable $X$ has a domain $dom(X)$ with cardinality $|X|$. For instance, $smokes(X)$ denotes an atomic formula over logical variable $X$, where, in our running example, the domain of $X$ contains Alice and Bob. $LV(\alpha)$ is the set of logical variables referred by $\alpha$, where $\alpha$ is a formula or a set of formulas.

A *substitution* $\theta$ over a set of logical variables $L$ maps each variable in $L$ to a constant symbol or some other logical variable, and $\alpha\theta$ is the result of applying $\theta$ to $\alpha$. A *ground substitution* is the substitution of all logical variables in $\alpha$ with constants. For instance, a substitution $Y/X$ applied to $friends(X, Y)$ results in $friends(X, X)$, and a ground substitution $X/Alice$ applied to atom $smokes(X)$ results in the ground atom $smokes(Alice)$. The range of atom $\alpha$, denoted by $range(\alpha)$, is defined as the range of any ground atom $\alpha\theta$ obtained by the ground substitution $\theta$.

A *constraint* $C$ is a set of formulas over a set of logical variables $L$, defining the set of legal substitutions applicable to $L$. For instance, the constraint $X \neq Y$ defines any substitution under which the substitutions of $X$ and $Y$ are identical, as illegal. We use $rv(\alpha : C)$ to depict the set of random variables that can be obtained by any legal substitution on $\alpha$, as defined by $C$.

$gr(L : C)$ is the set of legal ground substitutions which can be applied on $L$ under constraint $C$, where $|L : C|$ is used to depict the size of the set. We use $C_L^{\downarrow}$ to depict the projection of a set of logical variables $L$ on a constraint $C$. Similarly to [9], we require the constraints in our model to be in some normal form, where for each logical variable $X$, $|X : C|$ has a fixed value regardless of the binding of other logical variables in $C$.

Lastly, a *parfactor* is a tuple $(L, C, A, \phi)$, comprised of a set of logical variables $L$, a constraint $C$ on $L$, a set of atoms $A$ and a potential function $\phi : \prod_{\alpha \in A} range(\alpha) \to \mathbb{R}_0^+$. Applying substitution $\theta$ on parfactor $g = (L, C, A, \phi)$ results in $g\theta = (L', C\theta, A\theta, \phi)$, where $L'$ is obtained by applying a substitution on its logical variables, and dropping those who are mapped to constants. Parfactors compactly model a set of factors upon their *grounding*, namely – upon applying all legal ground substitutions, as defined by $C$. The operator $gr(g)$ depicts the set of all legal ground substitutions . The abbreviation $\phi : \alpha_1, \ldots, \alpha_m \mid C_\psi$ is used to describe a parfactor with the set of atoms $\bigcup_i \{\alpha_i\}$, a constraint with formula $\psi$, and an explicit set of logical variables.

**Example 2.2.** Let parfactor $g$ consist of the following, $\phi : smokes(X), drinks(Y), friends(X, Y) \mid C_{X \neq Y}$. In a world where the domain of $X$ and $Y$ equals

{Alice, Bob}, the grounding of $g$ entails two factors:
$\phi: smokes(Alice), drinks(Bob), friends(Alice, Bob)$
$\phi: smokes(Bob), drinks(Alice), friends(Bob, Alice)$

The weight of $g$ is defined as $w_g(v) = \prod_{f \in gr(g)} w_f(v)$, where $v$ is an assignment to all the ground atoms (random variables) entailed by the the grounding of $g$. This set of random variables is conveniently denoted by $rv(g)$.

### 2.2 Shattering, Fusion and Propositionalization

Let model $G$ be a set of parfactors, let $\alpha_i$ denote an atom, and $C_i$ denote its parfactor constraint. We say that $G$ is *completely shattered*, if for any $i, j$, the sets of random variables denoted by $rv(\alpha_i : C_i)$ and $rv(\alpha_j : C_j)$ are either disjoint or equal. Hence, *shattering* [2] is a procedure which takes model $G$, and produces a completely shattered model $G'$, s.t. the set of all ground factors entailed by $G$ and $G'$ is equal. To eliminate some possible ambiguity, we refer to all atoms in a completely shattered model as each having a unique predicate symbol.

**Example 2.3.** Let model $G$ consist of two parfactors, $\phi_1: p(X, Y), q(Y)$ and $\phi_2: p(X, X)$.
The shattering of $G$ replaces the first parfactor with two new parfactors, consisting of the same potential, as follows $\phi_1: p(X, X), q(X)$ and $\phi_1: p(X, Y), q(Y) \mid C_{X \neq Y}$.

*Fusion* [2] is the procedure of merging two or more parfactors by unifying logical variables, while maintaining the same model weight under any given assignment. Since there are several ways to unify logical variables, inference engines apply fusion according to some strategy. For instance, consider the fusion of $\phi_1: p(X, Y), q(Z)$ and $\phi_2: p(X, Y), q(Y)$. Two possible fusions (out of many) are $\phi_{f_1}: p(X, Y), q(Z), q(Y)$ and $\phi_{f_2}: p(X, Y), p(X, Z), q(Y)$, and so the inference engine must choose which one to apply. In this paper, fusion is used only symbolically (aka *symbolic fusion*), as a means of structure analysis. As a result, table entries are never examined, and the fusion procedure never "blows up".

*Propositionalization* [9] is the operation of partially grounding a parfactor over a set of logical variables, thereby eliminating these variables from the resulting set. For example, the propositionalization of $Z$ in parfactor $\phi: p(Y, Z), q(X, Y)$, where $dom(Z) = \{z_1, z_2\}$, results in the two parfactors, $\phi: p(Y, z_1), q(X, Y)$ and $\phi: p(Y, z_2), q(X, Y)$.

### 2.3 First-Order MPE

We now define the MPE task. Given a set of parfactors $G$, the combined weight of an assignment $v$ is

$$w_G(v) = \prod_{g \in G} w_g(v) = \prod_{g \in G} \prod_{f \in gr(g)} w_f(v) \quad (1)$$

The Most Probable Explanation of a first-order model is the pair $(v, \frac{1}{Z} w_G(v))$, s.t. $v = \arg\max_{v'} w_G(v')$, and $Z$ is a normalization factor. In this paper, as in [3], we address the task of obtaining the unnormalized MPE.

## 3 Walk-through: Simplifying the MPE Task

### 3.1 Motivation

Consider the following single-parfactor model:

$$\phi: friends(Y, X), friends(Z, X), knows(Y, Z)$$

In this model, the chance of two people knowing each other depends on having mutual friends. Although a simple model, all existing methods fail to lift the task of computing its partition function. More specifically, inversion elimination [2] fails since no atom contains all the logical variables. Counting conversion [2] and partial inversion [3] fail since all logical variables are occupied by at least two atoms, but never by all three. Recursive conditioning (splitting on atoms) [6] fails, since all atoms contain more than a single logical variable. Finally, additional elimination and model restructuring rules [7; 1] fail to assist as well.

However, obtaining the MPE of this same model, as we will demonstrate, is polynomial in the population size. This, of course, cannot be achieved by a simple modification of the lifted methods, but rather – by exploiting a property which is unique to MPE: *uniform assignments*. To understand this property and the derived framework, we shall study three model prototypes: *single-parfactor with no recurring formulas* (namely, each atomic formula appears only once), *single-parfactor with recurring formulas* and *multiple-parfactors*.

#### 3.1.1 Single Parfactor, No Recurring Formulas

Consider model $\phi: a(X), b(Y), c(Z)$. Computing the partition function of this model is polynomial in the domain size, but finding the MPE is even easier: let entry $\phi(\rho_a, \rho_b, \rho_c)$ be the maximum entry in table $\phi$. An assignment with maximal potential is obtained by assigning all grounds of atoms of $a(X), b(Y)$ and $c(Z)$ uniformly, with $\rho_a, \rho_b$ and $\rho_c$, respectively. We refer to $a(X), b(Y)$ and $c(Z)$ as *uniformly assigned formulas*.

We wish to exploit the UA property in a way that may seem redundant at the moment, by applying a procedure called *uniform assignment reduction* (UAR). UAR modifies the model while preserving the MPE probability, producing a simpler MPE task. In our example $\phi: a(X), b(Y), c(Z)$ is modified to $\phi': a', b', c'$, where the arity-reduced atoms $a', b', c'$ have the same assignment range as their original counterparts, and $\phi' = \phi^{|\{X,Y,Z\}|}$. This exponentiation compensates for the reduced number of ground factors post modification. Once the MPE of the modified model is obtained (by a simple table lookup), the MPE assignments for $a(X), b(Y)$ and $c(Z)$ are immediately derived.

### 3.1.2 Single Parfactor, Recurring Formulas

We return to the friends/knows example discussed earlier, $\phi : \ fr(Y,X), fr(Z,X), k(Y,Z)$ ( where $fr \equiv friends$, $k \equiv knows$). Unlike the previous case, a simple table lookup cannot resolve this model's MPE, since any assignment to one ground of $fr$ inevitably affects table entries in two table positions: $fr(Y,X)$ and $fr(Z,X)$. As a first step, we identify $\{Y,Z\}$ as the *overlap set* of the model, as $Y$ and $Z$ are different logical variables occupying the same position in predicate $fr$. Intuitively, the overlap set is the "obstruction" under which the MPE task cannot be solved by a simple table lookup. Hence, the next logical step would be to eliminate this obstacle.

Assume now that we propositionalize both $Y$ and $Z$, where the domain of both is $\bigcup_i \{o_i\}$. The result of this operation is a set of parfactors, residing over the set of atoms $A = \bigcup_{i,j} \{fr(o_i, X), k(o_i, o_j)\}$. If we fuse all these parfactors together without applying a "name change" to any of the logical variables, we obtain a single parfactor over $A$ with no recurring formulas. Thus, the previous case applies and each of $fr(o_1, X), \ldots, fr(o_n, X)$ is detected as uniformly assigned. That is, for $o_1$ and for *every* possible value of $X$, $fr(o_1, X)$ will have the same value assigned in an MPE. Similarly, for $o_2$ and for *every* possible value of $X$, $fr(o_2, X)$ will have the same value assigned in an MPE, etc. Of course, this maximizing value for $fr(o_i, X)$ and $fr(o_j, X)$ may be different, if $i \neq j$. Interestingly, we did not refer to any of the table entries in our procedure. In other words, the UA property can be detected by a set of purely symbolic operations.

We encapsulate the above sequence of operations within a new operator called *anchoring*, denoted by $\circledast$, which applies a set of symbolic propositionalizations and fusions. In our example, $\phi : \ fr(Y,X), fr(Z,X), k(Y,Z) \circledast \{Y,Z\} = \phi_f : \ fr(*,X), k(*,*)$, where $fr(*,X)$ and $k(*,*)$ are called *anchored formulas*: $fr(*,X) \equiv \bigcup_i \{fr(o_i, X)\}$ and $k(*,*) \equiv \bigcup_{i,j} \{k(o_i, o_j)\}$. We emphasize that anchoring is a symbolic operation that does not require actually carrying out explicit propositionalization and fusion but is equivalent semantically. Thus, anchoring allows us to analyze the model's structure regardless of domain size, and to simplify the depiction of uniform assignments. Namely, the anchored formula $fr(*, X)$ carries the information that $fr(o_1, X), \ldots, fr(o_n, X)$ are uniformly assigned. Essentially, this says that in the MPE, for every $o_j$, $fr(o_j, X)$ will have the same value for every possible substitution of $X$. Thus, it depends on $o_j$ alone, and we can use $fr'(o_j)$ to denote it, instead of $fr(o_j, X)$.

To sum up thus far, we identified an overlap set, applied anchoring and obtained a set of anchored formulas. As a last step, we wish to exploit the UA property implied by the anchored formulas. We return to the UAR procedure which was introduced previously. However, now arity reductions are applied discriminately, as only non-anchored logical variables (those remaining in the anchored formulas) can be removed from the model. Hence, only $X$ is eliminated by the UAR procedure, producing $\phi' : \ fr'(Y), fr'(Z), k(Y,Z)$, where $\phi' = \phi^{|X|}$. From here, the MPE is resolved by any existing inference engine.

### 3.1.3 Multiple Parfactors

Here, we study a model consisting of two parfactors $\phi_1 : \ p(X), p(Y), q(Z)$ and $\phi_2 : \ p(X), r(X), q(Z)$. As before, we wish to detect UAs and apply a UAR procedure. However, the UA detection in this case is somewhat more complicated, and involves a sequence of operations: anchoring, *alignment* and symbolic fusion. We start by identifying $\{X,Y\}$ as an overlap set in the first parfactor, followed by its anchoring to $\phi_{f_1} : \ p(*), q(Z)$. At this stage, the form of $p$ is not consistent among different parfactors in the model, since the second parfactor contains a non-anchored $p(X)$. To address this inconsistent form, we must *align* the model by anchoring the $X$ logical variable in the second parfactor. Once completed, the second parfactor is replaced with $\phi_{f_2} : \ p(*), r(*), q(Z)$.

The next step invokes a (symbolic) fusion of the two anchored parfactors (those consisting of $\phi_{f_1}$ and $\phi_{f_2}$), producing $\phi_f : \ p(*), r(*), q(Z)$. In general, symbolic fusions may produce additional overlaps, but in this simple case no overlaps are introduced. Hence, the anchored formulas now carry the UA property. Finally, UAR is applied independently on each of the original parfactors, producing $\phi_1^{|Z|} : \ p(X), p(Y), q'$ and $\phi_2^{|Z|} : \ p(X), r(X), q'$. Having introduced the core concepts of our algorithm, we move to a formal representation.

## 4 Formal Definitions

### 4.1 Overlap Set, Anchoring and Alignment

A *position* of logical variable $X$ in formula $\alpha$ is the location of $X$ in the formula's predicate under argument list ordering. In $p(X, Y)$, the positions of $X$ and $Y$ are 1 and 2 respectively.

An *overlap set* $L_o$ in parfactor $g$ is the set of all logical variables in $g$ which do not occupy the same positions under all instances of the same formulas.

**Example 4.1.** : In parfactor
$\phi : \ p(X,Z,Y), p(Z,U,Y), q(S,W), q(S,T), r(S)$,
the overlap set is $L_o = \{X, Z, U, W, T\}$.

The *effect scope* of a set of logical variables $L_o$ in parfactor $g = (L, C, A, \phi)$, denoted by $L_{eff}$, is the set of logical variables in $g$ whose set of substitutions is dependent on any binding of $L_o$. Formally, $L_{eff}$ is defined as the minimal set of logical variables in $g$ under which $L_o \subseteq L_{eff}$, and $\forall \theta_1, \theta_2 \in gr(L_{eff} : C) : gr(L \setminus L_{eff} : C\theta_1) = gr(L \setminus L_{eff} : C\theta_2)$. **In constraint-less parfactors, it is always the case that $L_{eff} = L_o$.**

**Example 4.2.** The effect scope of $\{X, W\}$ in parfactor $\phi : p(X), q(Y, Z), r(W)$ is $\{X, W\}$.

**Example 4.3.** The effect scope of $\{X\}$ in parfactor $\phi : p(X), q(Y), r(Z), s(W)|C_{X \neq Y \wedge Y \neq Z}$ is $\{X, Y, Z\}$.

*Anchoring* of parfactor $g = (L, C, A, \phi)$ over the set of logical variables $L_o$, denoted by $g \circledast L_o$, is a two step procedure: propositionalization of $L_{eff}$, which is the effect scope of $L_o$, followed by a symbolic fusion of the result set of parfactors. The final result, parfactor $g' = (L', C', A', \phi')$, contains the following properties:

- $L' = L \setminus L_{eff}$
- $C' = C_{L'}^{\downarrow}$
- $A' = \{\alpha\theta \mid \alpha \in A, \theta \in gr(L_{eff} : C)\}$

And some $\phi'$, which we refrain from actually computing in this context. The positions of the propositionalized logical variables under anchoring are called *anchored positions*. We use the substitution $\theta = \bigcup_{X \in L_{eff}} \{X/*\}$ on the set of logical variables in $g$ to depict the result of anchoring.

*Alignment* of parfactor $g$ according to parfactor $g_*$, denoted by $align(g, g_*)$, is the anchoring of $g$ over $L$, where $L$ consists of all logical variables in $g$ that occupy positions which are anchored in $g_*$, under predicates which are mutual to $g$ and $g_*$. For instance, aligning $\phi : p(X,Y), q(Z)$ according to $\phi_* : r(*), p(W,*)$, yields $\phi' : p(X,*), q(Z)$. We define $align(G, g_*)$ as applying $align(g, g_*)$ to all $g \in G$, and then repeatedly re-aligning the new content, until all instances of each predicate are anchored consistently.

### 4.2 Uniform Assignments and Arity Reduction

A *uniform assignment* over formula $\alpha$ is the assignment of a constant $\rho$ to every ground of $\alpha$, where $\rho \in range(\alpha)$. A *partially uniform assignment* over formula $\alpha$ with relation to a set of logical variables $L$ is a uniform assignment over each ground substitution of $L$. Namely, a partially uniform assignment over $t(X, Y, Z)$ w.r.t. $\{Y, Z\}$ implies that $\forall y \in dom(Y), \forall z \in dom(Z) : t(X, y, z)$ is uniformly assigned.

To enable the application of uniform assignments on our model, we define the *arity reduction* operator $\alpha \downarrow \alpha_*$ on atomic formulas. If $\alpha$ and $\alpha_*$ have the same predicate symbol, $\alpha \downarrow \alpha_*$ is a new predicate obtained by removing all positions in $\alpha$ not anchored in $\alpha_*$. For instance, $p(X, Y, Z) \downarrow p(W, *, T) = p'(Y)$. If both formulas differ in the predicate symbol, $\alpha \downarrow \alpha_* = \alpha$.

### 4.3 Uniform Assignment Reduction

A *Uniform Assignment Reduction* (*UAR*) applied to parfactor $g = (L, C, A, \phi)$ by formula $\alpha_*$, denoted by $g \downarrow \alpha_*$, is the operation of applying $\alpha \downarrow \alpha_*$ to each $\alpha \in A$, followed by exponentiating the potential by the combined domain size of the removed logical variables. More formally, the operation $g \downarrow \alpha_*$ yields $g' = (L', C', A', \phi')$ with the following properties:

- $A' = \{\alpha \downarrow \alpha_* \mid \alpha \in A\}$
- $C' = C_{L'}^{\downarrow}$
- $L' = LV(A')$
- $\phi' = \phi^{|(L \setminus L') : C\theta'|}$

Where $\theta'$ is some ground substitution over $L'$ under constraint $C$.

**Example 4.4.** Let $g = \phi : p(X), q(Y, Z)$. Let $\alpha_* = q(W, *)$. Hence, $g \downarrow \alpha_* = \phi^{|Y|} : p(X), q'(Z)$.

**Example 4.5.** Let $g = \phi : p(X), q(Y) \mid C_{X \neq Y}$. Let $\alpha_* = p(Z)$. Hence, $g \downarrow \alpha_* = \phi^{|X|-1} : p', q(Y)$, since $|X : C_{X \neq Y}| = |X| - 1$ under any binding of $Y$.

**Example 4.6.** When the set of the logical variables remains the same, as in $\phi : p(Y), q(Y, Z) \downarrow q(W, *) = \phi : p(Y), q'(Z)$, no exponentiation is required.

## 5 Simplified First-Order MPE

The model simplification algorithm is comprised of two phases – detecting uniform assignments, and applying a uniform assignment reduction, as follows.

---

**Algorithm 1:** DETECT-UNIFORM-ASSIGNMENTS

**input** : $G_{input}$ – a completely shattered model
**output**: A set of anchored formulas

$G \leftarrow G_{input}$
**while** $\exists g \in G, \exists L_o \neq \emptyset$ s.t. $L_o$ is an overlap set in $g$
**do**
    $g_* \leftarrow g \circledast L_o$     // anchoring
    $G \leftarrow align(G, g_*)$     // alignment

**if** $\exists g_1 \neq g_2 \in G$ s.t. $rv(g_1) \cap rv(g_2) \neq \emptyset$ **then**
    $G \leftarrow G \cup \text{SymbolicFusion}(g_1, g_2) \setminus \{g_1, g_2\}$
    **return**
    DETECT-UNIFORM-ASSIGNMENTS($G$)
**else**
    **return** all formulas in $G$

---

### 5.1 Detecting Uniform Assignments

The detection procedure (Algorithm 1) is purely symbolic, hence only the set of formulas is examined, whereas table entries are completely ignored. The detection starts by identifying all overlap sets in the model, accommodated by anchoring. The result is a set of parfactors with no recurring formulas. The entire model is then aligned accord-

**Algorithm 2:** SIMPLIFY-FOMPE

**input** : $G$ – a completely shattered model
**output**: Simplified Model

$A_* \leftarrow$ DETECT-UNIFORM-ASSIGNMENTS($G$)
**foreach** $\alpha_* \in A_*$ **do**
    **foreach** $g \in G$ **do** $G \leftarrow G \cup \{g \downarrow \alpha_*\} \setminus \{g\}$
**return** $G$

ingly, making sure that following manipulations operate on a completely shattered model. Next, two parfactors are chosen for a symbolic fusion. This, in turn, may generate a new overlap set. The cycle of overlap set detection, anchoring, and symbolic fusion is repeated until no two parfactors which share a formula are left. The procedure returns a set of anchored formulas, depicting the set of detected UAs.

One issue which remains unsolved is how to reapply the symbolic fusion procedure in order to produce as many arity reductions as possible. We use a greedy fusion scheme, where the logical variables of formulas with maximum arity are matched first. This, of course, does not guarantee an optimal sequence of fusions, and we leave the study of better fusion selection strategies for future work.

**Proposition 1.** *Given a completely shattered model $G$, Algorithm 1 produces a set of anchored formulas $A_*$, where there exists an MPE solution under which all formulas $\alpha_* \in A_*$ are partially uniformly assigned w.r.t. the logical variables in the anchored positions.*

*Proof outline.* The algorithm applies (symbolically) standard lifted inference tools, where each fusion is followed by anchoring, namely – the elimination of overlapping logical variables. Once all parfactors are merged into a single parfactor with no overlap sets, an MPE solution can be trivially obtained by assigning the atomic formulas uniformly, according to remaining logical variables. □

### 5.2 UAR Algorithm

After invoking DETECT-UNIFORM-ASSIGNMENTS, which detects uniform assignments in model $G$, a uniform assignment reduction is applied to each parfactor in $G$, producing a simplified model $G'$.

**Proposition 2.** *Algorithm 2 produces a simplified model $G'$ with the same MPE probability as $G$.*

*Proof outline.* The formal proof is found in the Appendix. The idea is to pair each original parfactor $g \in G$ with its UAR result $g' \in G'$, and show the following: (i) for each assignment $v'$ in model $G'$ there exists an assignment $v$ in model $G$, where the probabilistic weights of all pairs $g$ and $g'$ are equal; and (ii) for each optimal assignment $v$ in $G$, under which formulas are uniformly assigned according to $A_*$, there exists an assignment $v'$ s.t. the probabilistic weights of all pairs $g$ and $g'$ are equal. $A_*$ here denotes the set of anchored formulas which yielded $G'$. □

| | |
|---|---|
| Original model | $\phi_1 : p(X,Y), q(Y,Z), r(Z,X), s(X,Z)$<br>$\phi_2 : s(X,\underbrace{V}_{overlap}), s(X,\underbrace{W}_{overlap}), p(X,Y)$ |
| Anchoring | $\phi_1 : p(X,Y), q(Y,Z), r(Z,X), s(X,Z)$<br>$\phi'_2 : s(X,*), p(X,Y)$ |
| Model alignment | $\phi'_1 : p(X,Y), q(Y,*), r(*,X), s(X,*)$<br>$\phi'_2 : s(X,*), p(X,Y)$ |
| Symbolic fusion | $\phi_f : p(X,Y), q(Y,*), r(*,X), s(X,*)$ |
| UA reduction | $\phi_1^{|X| \cdot |Y|} : p', q'(Z), r'(Z), s'(Z)$<br>$\phi_2^{|X| \cdot |Y|} : s'(V), s'(W), p'$ |

Figure 1: Example – model simplification

#### 5.2.1 UAR Example

We now cover the example in Figure 1. Let model $G$ consist of $\phi_1 : p(X,Y), q(Y,Z), r(Z,X), s(X,Z)$ and $\phi_2 : s(X,V), s(X,W), p(X,Y)$. We start by detecting the overlap set {V,W} in $\phi_2$, followed by a subsequent anchoring of $\phi_2$ to $\phi'_2 : s(X,*), p(X,Y)$, and the alignment of $\phi_1$ to $\phi'_1 : p(X,Y), q(Y,*), r(*,X), s(X,*)$. Next, a symbolic fusion is applied to $\phi'_1$ and $\phi'_2$, resulting in $\phi_f : p(X,Y), q(Y,*), r(*,X), s(X,*)$. The remaining formulas in $\phi_f$ are then passed to the SIMPLIFY-FOMPE procedure, where consecutive uniform assignment reductions yield $\phi_1^{|X| \cdot |Y|} : p', q'(Z), r'(Z), s'(Z)$ and $\phi_2^{|X| \cdot |Y|} : s'(V), s'(W), p'$.

### 5.3 Complexity Analysis

Let $t$, $n$, $a$ and $p$ denote the properties of the original model $G$, where $t$ is the number of table entries, $n$ is the number of atoms, $a$ is the maximum arity of any predicate and $p$ is the number of parfactors. The complexity of SIMPLIFY-FOMPE is simply $O(t)$. As for DETECT-UNIFORM-ASSIGNMENTS, each position in any atom of $G$ is aligned no more than once, thus $O(n \cdot a)$ is the upper bound for the number of anchored positions. $p - 1$ is the maximal number of symbolic fusions. Combined, the total complexity is $O(t) + O(n \cdot a) + O(p \cdot O_f)$, where $O_f$ is the upper bound of a symbolic fusion and depends on the fusion strategy. For example, a greedy fusion strategy takes $O(n \cdot a)$ time. Other polynomial time strategies are possible. In typical models, $n$ and $t$ are the dominating elements, and so we get a complexity that is linear in $t$ and polynomial in $n$. All in all, the complexity is independent of domain sizes.

## 6 UA under Recursive Conditioning

As demonstrated, the UA property is heavily dependent on the structure of the model. One interesting aspect of recur-sive conditioning is the structure modification induced by the conditioning operator (splitting on a singleton atom). This allows non UA formulas to be split into uniformly assigned groups, a property which is referred to as a *conditional UA*. Consequently, the scope of uniform assignments is extended beyond the role of preprocessing, as presented so far, since various conditioning contexts introduce further opportunities for exploiting the UA property.

Consider model $\phi_1 : p(X), q(X)$ and $\phi_2 : p(X), q(Y)$, where $p$ and $q$ are boolean, and the domain size of $X$ and $Y$ is $n$. Since all possible fusions of $\phi_1$ and $\phi_2$ result in an overlap (e.g. $\phi_f : p(X), q(X), q(Y)$ ), this model contains no uniform assignments. However, when conditioning on atom $p(X)$, the ground atoms of $q(X)$ split into two uniformly assigned groups. We demonstrate this as follows: let $k \in \{0,...,n\}$ depict the number of $p(X)$ ground atoms assigned 0. Hence, under each binding of $k$, the domain is split into two parts, and a set of four logical variables is introduced: $X_0^k, X_1^k, Y_0^k$ and $Y_1^k$, where $|X_0^k| = k$, $|X_1^k| = n-k$ and $dom(X_i^k) = dom(Y_i^k)$. By definition, each ground of $p(X_i^k)$ is assigned $i$, thus all random variables of predicate $p$ are eliminated from the expression.

Formally:

$$\max_{rv(p(X),q(X))} \prod_X \phi_1(p(X),q(X)) \cdot \prod_{X,Y} \phi_2(p(X),q(Y)) =$$
$$\max_k \max_{rv(q(X))} \prod_i \prod_{X_i^k} \underbrace{\phi_1(i,q(X_i^k))}_{\equiv \phi_i^X(q(X_i^k))} \cdot$$
$$\prod_i \prod_{Y_i^k} \underbrace{\phi_2(0,q(Y_i^k))^k \phi_2(1,q(Y_i^k))^{n-k}}_{\equiv \phi_i^Y(q(Y_i^k))}$$

The result is a set of four parfactors, as follows:
(i) $\phi_0^X : q(X_0^k)$    (ii) $\phi_1^X : q(X_1^k)$
(iii) $\phi_0^Y : q(Y_0^k)$    (iv) $\phi_1^Y : q(Y_1^k)$

In this formation, it is easy to detect that both $q(X_0^k)$ and $q(X_1^k)$ are uniformly assigned. Namely, the set of parfactors can be simplified into
(i) $(\phi_0^X)^k : q_0'$    (ii) $(\phi_1^X)^{n-k} : q_1'$
(iii) $(\phi_0^Y)^k : q_0'$    (iv) $(\phi_1^Y)^{n-k} : q_1'$

We note that in this basic example, MPE can be efficiently solved without resorting to UAR. However, the same conditional UA principle applies to any model, thus the symbolic simplification becomes quite efficient when considering that the same computation is repeated for each $k$. Embedding UAR in a recursive conditioning inference engine, may indeed accelerate many of the lifted MPE tasks.

## 7 Experimental Evaluation

We present six sets of experiments. To highlight the fact that the method is independent of the inference engine used, and effective with different engines, we used two lifted inference engines, C-FOVE [9] and WFOMC [4], with and without the uniform assignment reduction. The engines were obtained from http://people.csail.mit.edu/milch/blog/index.html and dtai.cs.kuleuven.be/ml/systems/wfomc respectively, and were both modified to support max-product queries, by replacing Sum operators with Max, and ignoring the number of counting permutations. In WFOMC, specifically, the modification corresponds to changing the code in each of the NNF node subclasses. Times are shown in log scale, and the computation time of UAR was added to the overall time results.

Since WFOMC accepts weighted first-order formulas as input, each parfactor $\phi : p_1(X_1),...,p_n(X_n)$ was passed to WFOMC as a weighted first-order formula of the form $p_1(X_1) \wedge ... \wedge p_{n-1}(X_{n-1}) \implies p_n(X_n)$. Unlike variants of belief propagation [14; 8], where table entries affect the runtime of the algorithm, the runtime of both WFOMC and C-FOVE in models which exclusively consist of non-zero and non-one entries is agnostic to the actual numerical values. Hence, all table entries in the models presented here were arbitrarily set to non-zero, non-one.

Figures 2 & 3 show the results of two models that were previously discussed in this paper, demonstrating how UAR extends the scope of known tractable models. In these two figures, we see that both C-FOVE and WFOMC quickly fail to lift the inference task, as they resort to propositional inference. Once UAR is applied, both solve the MPE query efficiently. Figure 4 introduces a model which is solvable in polynomial time, and again, becomes much easier to solve with UAR. The ↓ symbol denotes a logical variable which was eliminated by the UAR.

In all three figures, UAR computation never exceeds 40 milliseconds. This result is, of course, consistent with the complexity analysis, demonstrating that UA detection has a negligible effect on the overall performance. Given that MPE is computationally demanding, and exponential in the worst-case, one could hardly lose from attempting to run UAR in every MPE task.

Figure 5 presents the results of running WFOMC with and without UAR, on random models with varying number of parfactors. Each parfactor, in an $n$-parfactor randomized model, consists of 3 atoms. Each atom is obtained randomly from a uniformly distributed pool of $2n$ unary predicates, and a set of (maximum) 3 logical variables per parfactor. This randomization guarantees a (polynomially) tractable model, that can evaluate the effectiveness of UAR. However, even with small domain sizes, solving MPE is still computationally demanding. We tested these random models under three domain sizes: 5, 10 and 50. The results are an average over 10 randomized sets.

Figure 6 shows an unsuccessful attempt by UAR to simplify the Friends & Smokers model [14]. Since UAR com-putation incurs very little overhead (48 milliseconds in this case), its effect on the result is negligible. In Figure 7, however, UAR is able to simplify the Students & Professors link prediction model [6], s.t. the computation becomes domain-size independent.

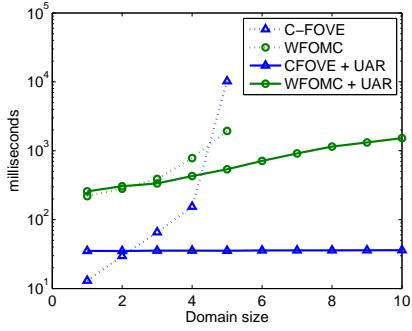
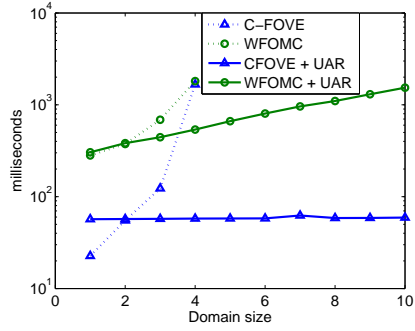
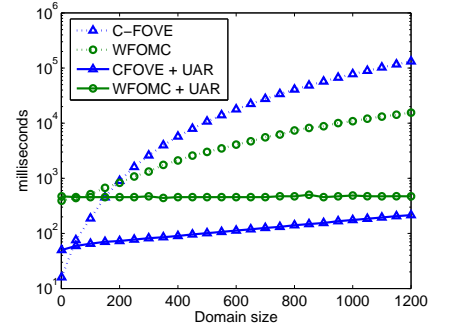

Figure 2: $\phi : friends(X_\downarrow, Y),$
$friends(X_\downarrow, Z), knows(Y, Z)$

Figure 3: $\phi_1 : p(X_\downarrow, Y_\downarrow), q(Y_\downarrow, Z),$
$r(Z, X_\downarrow), s(X_\downarrow, Z)$
$\phi_2 : s(X_\downarrow, V), s(X_\downarrow, W), p(X_\downarrow, Y_\downarrow)$

Figure 4: $\phi_1 : p(X), q(X), r(Z_\downarrow)$
$\phi_2 : p(X), q(Y), r(Z_\downarrow)$

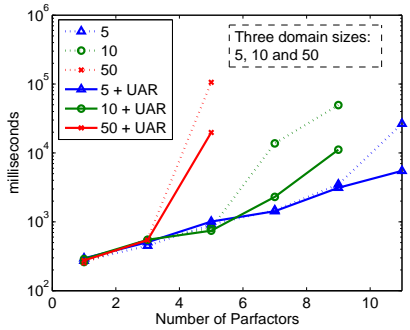
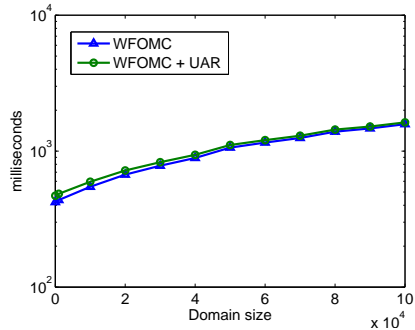
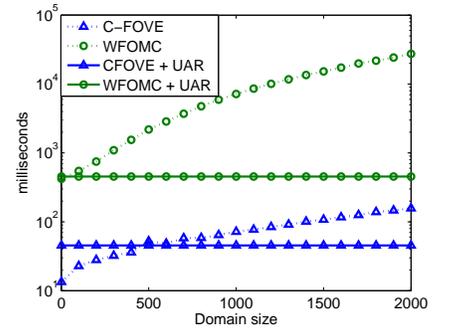

Figure 5: WFOMC on random models

Figure 6: Friends & Smokers [14]

Figure 7: Students & Professors [6]

## 8 Conclusion

We introduced a model simplification method which narrows the assignment space of first-order MPE queries by batching together sets of random variables, while refraining from table lookups. The suggested method can be integrated into existing lifted inference engines or serve as a preprocessing algorithm, adding a low computational overhead compared to non-simplified queries. And thus, even in cases where UAR fails, the penalty for running this procedure is negligible.

We note that the UA property is not restricted to the specific detection method presented in this paper. For example, although we refrained from analyzing the content of the potential tables when detecting uniform assignments, one can leverage specific maximizing assignments to obtain farther reductions. Consider, for example, an atom $p$ contained exclusively by parfactor $\phi : p(Y), p(Z)$. The overlap set of the two instances of $p$ is not empty, and in general requires anchoring and alignment. However, if maximal entries in $\phi$ assign both instances of $p$ the same value (e.g., they are $(0,0)$ or $(1,1)$, etc.) then $p$ can be treated as uniformly assigned. Namely, the parfactor is simplified into $\phi^{|Y|\cdot|Z|} : p'$, by applying UAR. This idea can be further adapted by approximation algorithms, which may force uniform assignments in cases where they are not guaranteed. Finally, third parties may also wish to assert uniform assignments as part of user constraints. For all these purposes, the UAR serves as a valid reduction.

### Acknowledgments

The authors are partially supported by ISF grant 8254320, the Paul Ivanier Center for Robotics Research and Production Management, and the Lynn and William Frankel Center for Computer Science.

## Appendix – UAR Proof

Let $G$ be a completely shattered model, and let $G' = \bigcup_{g \in G}\{g \downarrow \alpha_*\}$ be the set of post-UAR parfactors, for some anchored formula $\alpha_*$. WLOG, properties of each parfactor $g$ in $G$ are $(L, C, A, \phi)$, where $A = \{\alpha_1, \ldots, \alpha_n, \beta\}$, and $\forall i : rv(\alpha_i) = rv(\alpha_*) \wedge rv(\alpha_i) \cap rv(\beta) = \emptyset$.

Properties of each $g' = g \downarrow \alpha_* = (L', C', A', \phi')$ are

- $A' = \{\alpha'_1, \ldots, \alpha'_n, \beta\}$, where $\forall i : \alpha'_i = \alpha_i \downarrow \alpha_*$
- $L' = LV(A')$
- $C' = C^{\downarrow}_{L'}$
- $\phi' = \phi^{|(L \setminus L') : C\theta'|}$

Where $\theta'$ is some ground substitution over $L'$ under constraint $C$.

In the context of parfactor $g$, let $L_i^*$ denote the set of logical variables in $\alpha_i$ that occupy positions which are anchored in $\alpha_*$, and let $L_\beta$ depict the set of logical variables in $\beta$. We define $L^+ = L' = L_1^* \cup \ldots \cup L_n^* \cup L_\beta$, followed by $L^- = L \setminus L^+$ which is the set of logical variables which occupy positions that are unanchored in $\alpha_*$ and contained exclusively by some $\alpha_i$ instances. Note that $gr(L^+ : C) = gr(L' : C')$, since $L^+ = L'$ and $C' = C^{\downarrow}_{L'}$.

Weights of each $g$ and $g'$, under assignments $v$ and $v'$, are:

$$w_g(v) = \prod_{\theta^+} \prod_{\theta^-} \phi(\alpha_1 \theta^+ \theta^-_{|v}, \ldots, \alpha_n \theta^+ \theta^-_{|v}, \beta \theta^+ \theta^-_{|v})$$

$$w_{g'}(v') = \prod_{\theta'} \phi(\alpha'_1 \theta'_{|v'}, \ldots, \alpha'_n \theta'_{|v'}, \beta \theta'_{|v'})^{|L \setminus L' : C\theta'|}$$

Where $\prod_{\theta^+}$, $\prod_{\theta^-}$ and $\prod_{\theta'}$ are abbreviations for $\prod_{\theta^+ \in gr(L^+:C)}$, $\prod_{\theta^- \in gr(L^-:C\theta^+)}$ and $\prod_{\theta' \in gr(L':C')}$.

**Proposition 3.** *For each $v'$ there exists $v$ such that $w_G(v) = w_{G'}(v')$*

*Proof.* We pick one of the $\alpha_i$ instances in some parfactor $g$, and construct $v$ out of $v'$ as follows:

1. $\forall \theta^* \in gr(L_i^* : C)$, $\forall \theta \in gr(LV(\alpha_i) \setminus L_i^* : C\theta^*)$ : $\alpha_i \theta^* \theta_{|v} = \alpha'_i \theta^*_{|v'}$

2. $\psi \in rv(G) \wedge \psi \in rv(G') \implies \psi_{|v} = \psi_{|v'}$
   Meaning, the assignment to variables which are mutual to $G'$ and $G$ is copied from $v'$ to $v$.

The weights of each $g$ and counterpart $g'$ are then expressed as follows

$$w_g(v) = \prod_{\theta^+} \prod_{\theta^-} \phi(\alpha_1 \theta^+ \theta^-_{|v}, \ldots, \alpha_n \theta^+ \theta^-_{|v}, \beta \theta^+ \theta^-_{|v})$$

$$= \prod_{\theta^+} \prod_{\theta^-} \phi(\alpha_1 \theta^+ \theta^-_{|v'}, \ldots, \alpha_n \theta^+ \theta^-_{|v'}, \beta \theta^+ \theta^-_{|v'})$$

$$= \prod_{\theta^+} \prod_{\theta^-} \phi(\alpha'_1 \theta^+_{|v'}, \ldots, \alpha'_n \theta^+_{|v'}, \beta \theta^+_{|v'})$$

$$= \prod_{\theta'} \phi(\alpha'_1 \theta'_{|v'}, \ldots, \alpha'_n \theta'_{|v'}, \beta \theta'_{|v'})^{|L \setminus L' : C\theta'|}$$

$$= w_{g'}(v')$$

Since the full weights are a product over all parfactor weights, it follows that $w_G(v) = w_{G'}(v')$. $\square$

**Proposition 4.** *Given optimal assignment $v = \text{argmax}_v w_G(v)$, under which $\alpha_*$ is uniformly assigned, there exists $v'$ for which $w_{G'}(v') = w_G(v)$.*

*Proof.* First, we pick one of the $\alpha'_i$ instances in some parfactor $g'$. From the uniform assignment property depicted by $\alpha_*$, it follows that $\forall \theta^* \in gr(L_i^* : C)$, $\forall \theta_1, \theta_2 \in gr(LV(\alpha_i) \setminus L_i^* : C\theta^*) : \alpha_i \theta^* \theta_1(v) = \alpha_i \theta^* \theta_2(v)$. Next, we construct $v'$ out of $v$ as follows:

1. $\forall \theta^* \in gr(L_i^* : C)$, $\forall \theta \in gr(LV(\alpha_i) \setminus L_i^* : C\theta^*)$ : $\alpha'_i \theta^*_{|v'} = \alpha_i \theta^* \theta_{|v}$
   Enabled by the uniform assignment property.

2. $\psi \in rv(G) \wedge \psi \in rv(G') \implies \psi_{|v'} = \psi_{|v}$

As previously, weights of each $g'$ and $g$ are expressed as a function of ground substitutions, where $w_{g'}(v') = w_{g'}(v)$, and consequently $w_{G'}(v') = w_G(v) = \max_v w_G(v)$. $\square$

From propositions 3 & 4 it follows that $\max_{v'} w_{G'}(v') = \max_v w_G(v)$, hence the correctness of UAR.